\documentclass[journal]{IEEEtran}
\usepackage{graphicx}
\usepackage{enumerate}
\usepackage{amsmath,amsfonts,amssymb}
\usepackage{color}
\usepackage{multirow}
\usepackage[noend]{algpseudocode}
\usepackage{algorithmicx,algorithm}
\usepackage{setspace}
\usepackage{lineno,hyperref}
\usepackage{booktabs}
\usepackage[T1]{fontenc}
\usepackage[table]{xcolor}
\usepackage[numbers,sort&compress]{natbib}
\usepackage{xfrac}
\modulolinenumbers[5]
\usepackage{etoolbox}

\begin{document}

\title{SeaDATE: Remedy Dual-Attention Transformer with Semantic Alignment via Contrast Learning for Multimodal Object Detection}

\author{
Shuhan Dong, Yunsong Li,~\IEEEmembership{Member,~IEEE}, Weiying Xie, ~\IEEEmembership{Senior Member,~IEEE}, Jiaqing Zhang, Jiayuan Tian, Danian Yang, JieLei, ~\IEEEmembership{Member,~IEEE}

\thanks{This work was supported in part by the National Natural Science Foundation of China under Grant 62071360.  (\emph{Corresponding~authors: Yunsong Li; Weiying Xie.})

Shuhan Dong, Yunsong Li, Weiying xie, Jiaqing Zhang, Jiayuan Tian and Danian Yang are with the State Key Laboratory of Integrated Services Networks, Xidian University, Xi’an 710071,
China (e-mail: shdong@stu.xidian.edu.cn; ysli@mail.xidian.edu.cn; wyxie@xidian.edu.cn; jqzhang\underline{ }2@stu.xidian.edu.cn; jytian0414@stu.xidian.edu.cn; 1243313479@qq.com).

Jie Lei is with School of Electrical and Data Engineering, University of
Technology Sydney, Sydney, Australia (e-mail: jie.lei@uts.edu.au).

}
}

\markboth{JOURNAL OF LATEX CLASS FILES,~Vol.~X, No.~X, 2024}
{Dong \MakeLowercase{\textit{et al.}}: SeaDATE: Remedy Dual-Attention Transformer
with Semantic Alignment via Contrast Learning for Multimodal Object Detection}
\maketitle

\begin{abstract}  

Multimodal object detection leverages diverse modal information to enhance the accuracy and robustness of detectors. By learning long-term dependencies, Transformer can effectively integrate multimodal features in the feature extraction stage, which greatly improves the performance of multimodal object detection. However, current methods merely stack Transformer-guided fusion techniques without exploring their capability to extract features at various depth layers of network, thus limiting the improvements in detection performance. In this paper, we introduce an accurate and efficient object detection method named SeaDATE. Initially, we propose a novel dual attention Feature Fusion (DTF) module that, under Transformer's guidance, integrates local and global information through a dual attention mechanism, strengthening the fusion of modal features from orthogonal perspectives using spatial and channel tokens. Meanwhile, our theoretical analysis and empirical validation demonstrate that the Transformer-guided fusion method, treating images as sequences of pixels for fusion, performs better on shallow features' detail information compared to deep semantic information. To address this, we designed a contrastive learning (CL) module aimed at learning features of multimodal samples, remedying the shortcomings of Transformer-guided fusion in extracting deep semantic features, and effectively utilizing cross-modal information. Extensive experiments and ablation studies on the FLIR, LLVIP, and M$^{3}$FD datasets have proven our method to be effective, achieving state-of-the-art detection performance.

\end{abstract}
\section{Introduction}
\IEEEPARstart{O}{BJECT} detection \cite{survey}, an essential and challenging research task in computer vision, finds extensive applications in various fields such as image analysis \cite{analysis}, autonomous driving \cite{tracking1,tracking2}, and security surveillance \cite{security}. In recent decades, with the rapid progression of deep learning technologies, detection methods utilizing deep neural networks (DNNs) \cite{deeplearning} have witnessed significant improvements, leading to a substantial enhancement in detection capabilities.

\begin{figure}[htbp]
\centering
\includegraphics[width=0.48\textwidth]{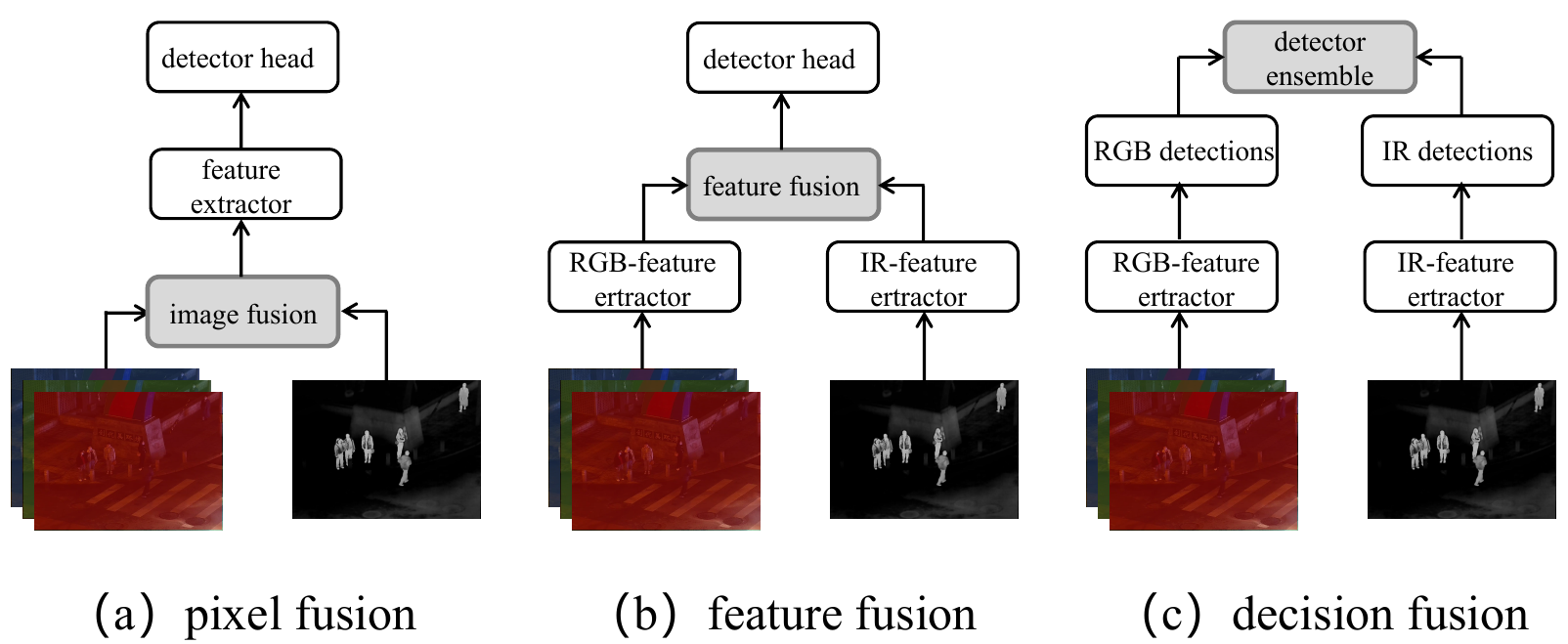}
\caption{Differernt strategies for fusion.}
\label{fig:1}
\end{figure}
Conventional visible object detection methods rely heavily on the rich textures and color details present in visible images. However, in difficult situations \cite{challengs,TCSVT3}, such as adverse weather, low light, or physical barriers, these methods often fail to maintain high accuracy in object detection task. In contrast, infrared sensors \cite{sensors}, which detect variations in thermal radiation, offer clearer outlines in such challenging conditions. Therefore, combining the complementary advantages of visible and infrared images through multimodal information fusion \cite{TCSVT1} substantially enhances the precision, dependability, and robustness of object detection algorithms.


Deep learning \cite{muldetection1,muldetection2,muldetection3}, renowned for its advanced feature extraction capabilities, has excelled in the realm of fusing visible and infrared images for object detection \cite{vis-inf}, thereby establishing itself as the principal technology in multimodal detection. Multimodal fusion, according to its stages of implementation \cite{trends}, can be categorized into three stages: pixel-level fusion, feature-level fusion, and decision-level fusion, as shown in Fig. \ref{fig:1}. Pixel-level fusion takes place prior to feature extraction but often encounters issues with information redundancy. Meanwhile, decision-level fusion combines the results of different modalities after model training. For instance, Chen \emph{et al.} \cite{chen} concentrated on post-optimization fusion, applying Bayes' rule and first principles to derive ProbEn, which assumes conditional independence between modalities. Li \emph{et al.} \cite{confidence} introduced a confidence perception mechanism that steers fusion processing via confidence estimation. Nevertheless, these methods tend to neglect interactions between different modalities, resulting in potential loss of information. In contrast, feature-level fusion occurs during the feature extraction phase. For example, Konig \emph{et al.} \cite{konig} developed a feature fusion region proposal network and experimentally identified the optimal layer for fusion. Ding \emph{et al.} \cite{ding} employed selective kernel units of varying sizes to merge features across layers using a feature pyramid. Most research in feature-level fusion relies on deep convolutional neural networks (CNNs) to devise more effective fusion methods, without addressing the inherent limitations of CNNs.

Although feature-level fusion demonstrates superiority over other methods, CNNs' limited local receptive field \cite{TCSVT2} and lack of long-term dependencies restrict their ability to fully utilize modal complementarity, owing to the convolution operator has a nonglobal receptive field. To address the inherent challenges of CNNs, Fang \emph{et al.} \cite{cft} proposed a cross-modal feature fusion method based on Transformer \cite{transformer}. Transformer's self-attention mechanism, with its global receptive field, is aptly suited for multimodal information fusion \cite{cnn-trans}. This method
treats the image as a pixel sequence and conduces fusion on pixel points. While it achieves excellent ability in detail information extraction, it overlooks the global interaction of information. Meanwhile, existing multimodal object detection methods apply Transformer-guided fusion simplistically at both shallow and deep networks without special processing that combines the essence of shallow and deep features. In fact, shallow features pay attention to details, and deep features pay attention to semantic content \cite{shallow-deep}. \textbf{Based on the above analysis, we hypothesize Transformer-guided fusion method may excel on shallow features, especially in capturing detailed information, as opposed to the semantic information of deep features.} In this paper, we explore the effectiveness of Transformer-guided fusion approach on different features at different locations, which is validated in Section \ref{Section 3}.


In this paper, we propose a novel multimodal object detection method using transformer to design a dual attention mechanism which includes spatial level and channel level. Spatial attention refines local representations by performing fine-grained interactions across spatial locations, and channel attention considers all spatial locations, naturally capturing global interactions and representations. This mechanism is integrated into the backbone network to enhance feature fusion through long-term dependencies. Moreover, we innovatively blend contrastive learning \cite{simclr} into multimodal object detection by analying the essence of shallow and deep features, facilitating deeper feature interplay. As a prominent self-supervised learning method, contrastive learning focuses on abstract semantic information and similarities among comparable instances. This synergy not only remedies the Transformer fusion mechanism but also probes into the complex interrelations between modalities, effectively utilizing cross-modal information.

The primary contributions of our research include:

\begin{itemize}
    \item 
We design a DTF feature fusion method, in which the dual attention mechanism takes into account both spatial level and channel level information interaction, and uses complementary modal information to effectively integrate multi-modal features.
    \item 

We design a CL module which integrated contrastive learning into the deepest layer of the network, facilitating complementarity and information exchange between deep features, remedying the shortcomings of Transformer fusion method in extracting semantic information.
    \item
Our approach significantly improves the accuracy of object detection. As tested on public datasets such as FLIR, LLVIP, and M$^{3}$FD, our method demonstrates superior detection performance compared to other leading technologies.
\end{itemize}

\section{Related Work}
\label{sec:Related Work}
\begin{figure*}[htbp]
\centering
\includegraphics[width=0.90\textwidth]{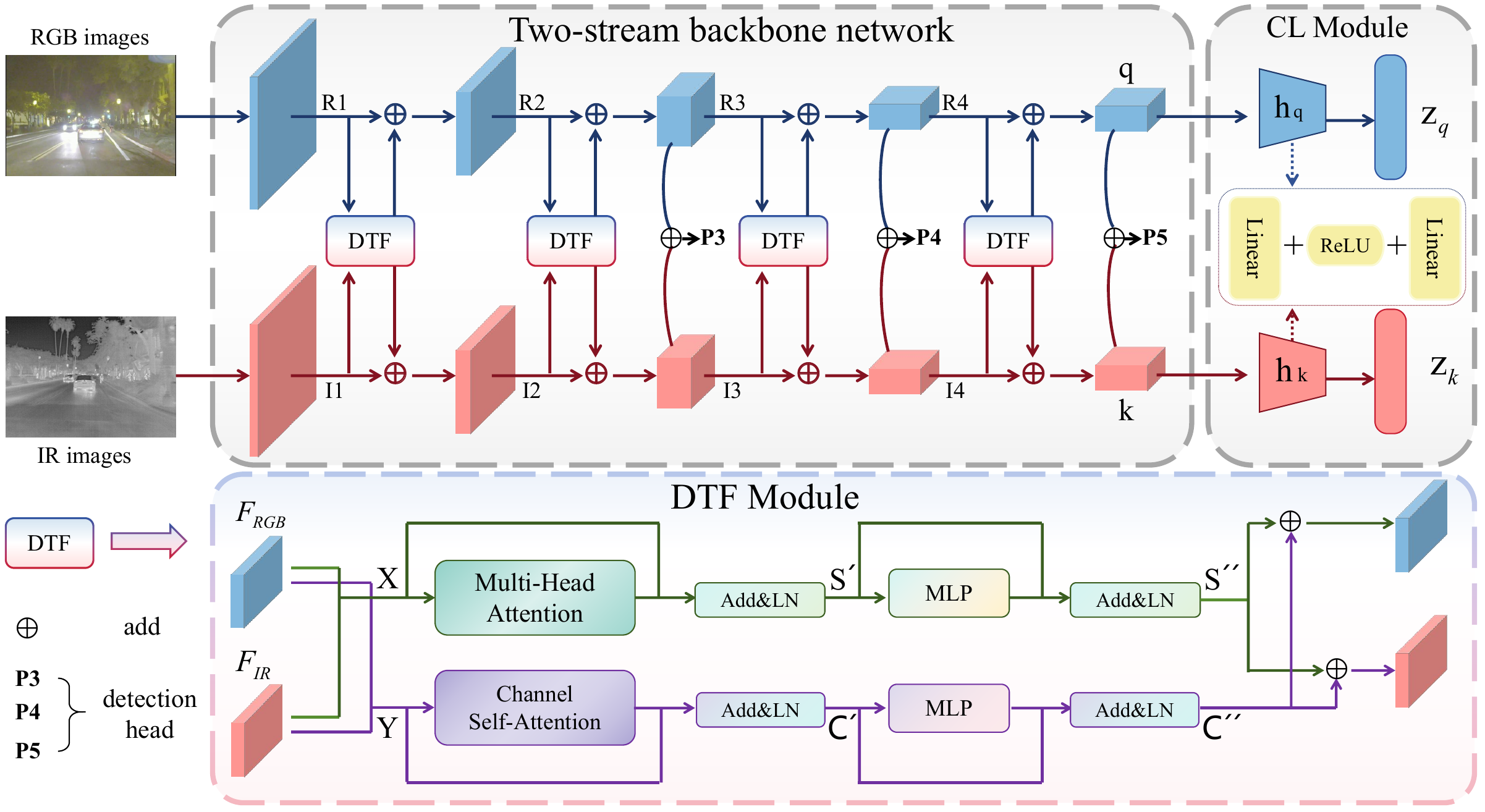}
\caption{Overview framework of the proposed SeaDATE. The network architecture is bifurcated into three components: 1) The feature extraction module with dual-stream E-ELAN as the backbone network. 2) DTF module for multi-modal feature fusion. 3) CL module for remedying Transformer fusion method.}
\label{fig:2}
\end{figure*}
\subsection{Multimodal Object Detection}
\label{Section 2.1}


The stability, reliability, and robustness of monomodality object detection are insufcient to deal with complex application environments. Multimodality object detection, such as visible–infrared fusion \cite{TCSVT4,TCSVT5}, can complementarily capture richer object information and obtain accuracy and more stable detection results. Since the process of multimodal detection is the same as monomodal detection, many related studies are based on the traditional RGB detection models such as the R-CNN series \cite{rcnn,fastrcnn,fasterrcnn,maskrcnn,rcfn}, and YOLO \cite{yolov1,yolov2,yolov3,yolov4}. Cao \emph{et al.} \cite{cao} adopted Faster R-CNN as the detection framework, in which additional backbone network was added as the feature extractor for the IR modality, which improved the performance of multimodal object detection through channel switching and spatial attention. For YOLO-MS \cite{yolo-ms}, the multimodal detector is based on YOLOv5. By designing a feature interaction module, the shared branch inside the module is used to enhance the interaction between the two modalities, breaking the isolation between the visible and infrared branches. Additionally, MBNet \cite{mbnet} utilize illumination-aware modules that allow the detection model to adjust weights of different input modalities based on light conditions, further improving the model’s performance.

\subsection{Fusion Strategies in Multimodal Object Detection}
\label{Section 2.2}
As the core problem of multimodal object detection, multimodal feature fusion are mainly divided into two research directions: one focusing on the choice of fusion stages and the other on the construction of fusion functions.

In studies of fusion stages, most efforts are dedicated to exploring the optimal fusion stage through the design of macro network architectures. Wanger \emph{et al.} \cite{Wagner} was the first to study the performance of pixel-level and decision-level fusion architectures on multimodal data. Liu \emph{et al.} \cite{liu} introduced feature fusion, proving its superior efficacy in multimodal object detection through the experiment, a finding further validated by subsequent studies \cite{research1,research2,research3}, indicating that feature fusion significantly outperforms other fusion approaches.

Regarding fusion functions, constructing an effective fusion mechanism is also crucial. The elementary approaches include direct merging of visible and infrared feature maps via concatenation, addition, maximal values, or cross-multiplication. Further advancements have progressed towards more sophisticated and logical fusion methods. For instance, Zhou \emph{et al.} proposed a novel channel-wise differential weighting method, utilizing the differential concept to suppress common-mode and amplify differential-mode, thereby improving multimodal feature fusion and achieving complementarity. Since the advent of the Transformer in 2017, with its straightforward yet potent architecture, it has rapidly expanded into computer vision and multimodality domains \cite{trans1,trans2,trans3}. For example, Yu \emph{et al.} \cite{yu} constructed a multimodal Transformer model, which achieved excellent results by capturing both intra-modal and inter-modal interactions in an unified attention block. 

Therefore, this paper proposes a Transformer-based feature fusion strategy, employing a dual attention mechanism to consider both spatial and channel information simultaneously for multimodal object detection task, aiming to further advance the development of this field.

\subsection{Contrastive Learning}
Self-supervision learning has become increasingly notable for its capability to autonomously uncover the inherent structure within data. Contrastive learning, as a key method in this domain, extensively utilized in areas like natural language processing and computer vision. This approach primarily operates at an abstract semantic level, discerning similarities among akin instances and differences between varied ones, thus demonstrating robust generalization potential. 

Initially, contrastive learning methods lacked a unified standard in model design and downstream task. Wu \emph{et al.} \cite{wu} treated each instance as a category, introducing individual discrimination tasks through feature differentiation, while Li \emph{et al.} \cite{li} used contrastive learning for effective predictions. The field saw a major shift with the introduction of MoCo \cite{moco} and SimCLR, which brought significant progress and innovation in the field.

In our research, the backbone of detection network serves as an encoder, offering deep features for contrastive learning task. Additionally, we adopt the queue dictionary proposed by MoCo-v2 \cite{moco2} and forgo momentum updates. The contrastive learning loss is incorporated into the primary network loss, optimizing and iterating the model parameters. This integration heightens the semantic similarity across different modalities, allowing the model to more effectively learn and complement multimodal information. Consequently, this leads to a notable enhancement in the model’s overall performance.


\section{Method}
\label{Section 3}

In this section, we delve into the details of SeaDATE network architecture. As depicted in Fig. \ref{fig:2}, we have developed a dual-stream backbone network, utilizing the YOLOv7 framework, specifically tailored for extracting features from both visible and infrared images for multimodal object detection. On one front, we've integrated a DTF module between these two modalities, enhancing the interplay of multimodal information and enabling effective multimodal fusion. On the other, we've embedded contrastive learning within the deeper layers of our feature extraction network. This integration is strategically designed to fully leverage the semantic information inherent in deep features. Following this, the combined features are fed into the detection head for subsequent analysis and processing.

\subsection{Dual-Attention Transformer Fusion Module}
\label{Section 3.1}

Firstly, we conducted ablation experiments on the FLIR datasets to test the previously stated hypothesis. As indicated in Table \ref{tbl:location}, the numbers 1, 2, 3, and 4 signify the placement of the DFT module within the dual-stream backbone network. A higher number corresponds to a deeper location of the DFT module within the network. We evaluated the model's detection performance at these varying positions using metrics such as $\text{mA}{{\text{P}}_{\text{50}}}$, $\text{mA}{{\text{P}}_{\text{75}}}$, and mAP. The results demonstrate that the Transformer-guided fusion method attains superior detection performance particularly in shallow features.  This suggests that the transformer-guided fusion approach is more adept at enhancing the details of shallow features compared to processing the semantic information of deeper features.

\begin{table}[tpb]
	\small
 \renewcommand{\arraystretch}{1.6}
	\centering
	\setlength{\tabcolsep}{3mm}{
		\caption{Ablation Experiment of DTF Module Position on FLIR Dataset}
		\label{tbl:location}
		\begin{tabular}{c|cccc}
			\toprule[1.2pt]
			\textbf{Location} 
                & \textbf{1}  & \textbf{2} & \textbf{3}  & \textbf{4}\\
			\midrule			 $\text{mA}{{\text{P}}_{\text{50}}}$ & \textbf{79.6}   & 78.6  & 78.4   & 78.4  \\
               $\text{mA}{{\text{P}}_{\text{75}}}$ &\textbf{35.0}   &34.7  &34.2 &34.1 \\
			$\text{mAP}$  &\textbf{40.6}    &39.8   & 38.8    & 38.4 \\

			\bottomrule[1.2pt]
	\end{tabular}}
\vspace{-0.1in}
\end{table}

Transformer are renowned for their adeptness at global context modeling, yet their computational complexity scales quadratically with the increase in token length.  Contemporary approaches predominantly utilize pixels or image patches as tokens, but this pixel-level self-attention leads to the loss of global information. Therefore, building on the traditional self-attention mechanism that facilitates pixel-level interactions among spatial tokens, this paper proposes a self-attention mechanism utilizing channel tokens. Each channel token is global in the spatial dimension, containing an abstract representation of the entire image. Consequently, the information exchange in channel self-attention occurs from a global perspective, enabling the capture of global contextual information while maintaining computational efficiency.

\subsubsection{Spatial Multi-Head Attention}
The goal of Spatial Multi-Head Attention is to facilitate information interaction at spatial locations using pixel-level tokens, capturing the complementary information between RGB and IR images. In this process, the spatial dimension $HW$ determines the count of tokens, while the channel dimension $C$ defines the feature size of tokens. Specifically, given the intermediate RGB features $F_{RGB} \in \mathbb{R} ^{C\times H\times W}$ and thermal feature maps$F_{IR} \in \mathbb{R} ^{C\times H\times W}$. Subsequently, these modality-specific sentences are concatenated to create the Transformer's input sentence $X_{RGB} \in \mathbb{R} ^{HW\times C}$ and $X_{IR} \in \mathbb{R} ^{HW\times C}$. Next, we concatenate the sentences from each modality to obtain the input sentence $X \in \mathbb{R} ^{2HW\times C}$. This input sentence S is projected onto three distinct weight matrices, generating sets of queries $Q$, keys $K$, and values $V$, 

\begin{figure}[htbp]
\centering
\includegraphics[width=0.48\textwidth]{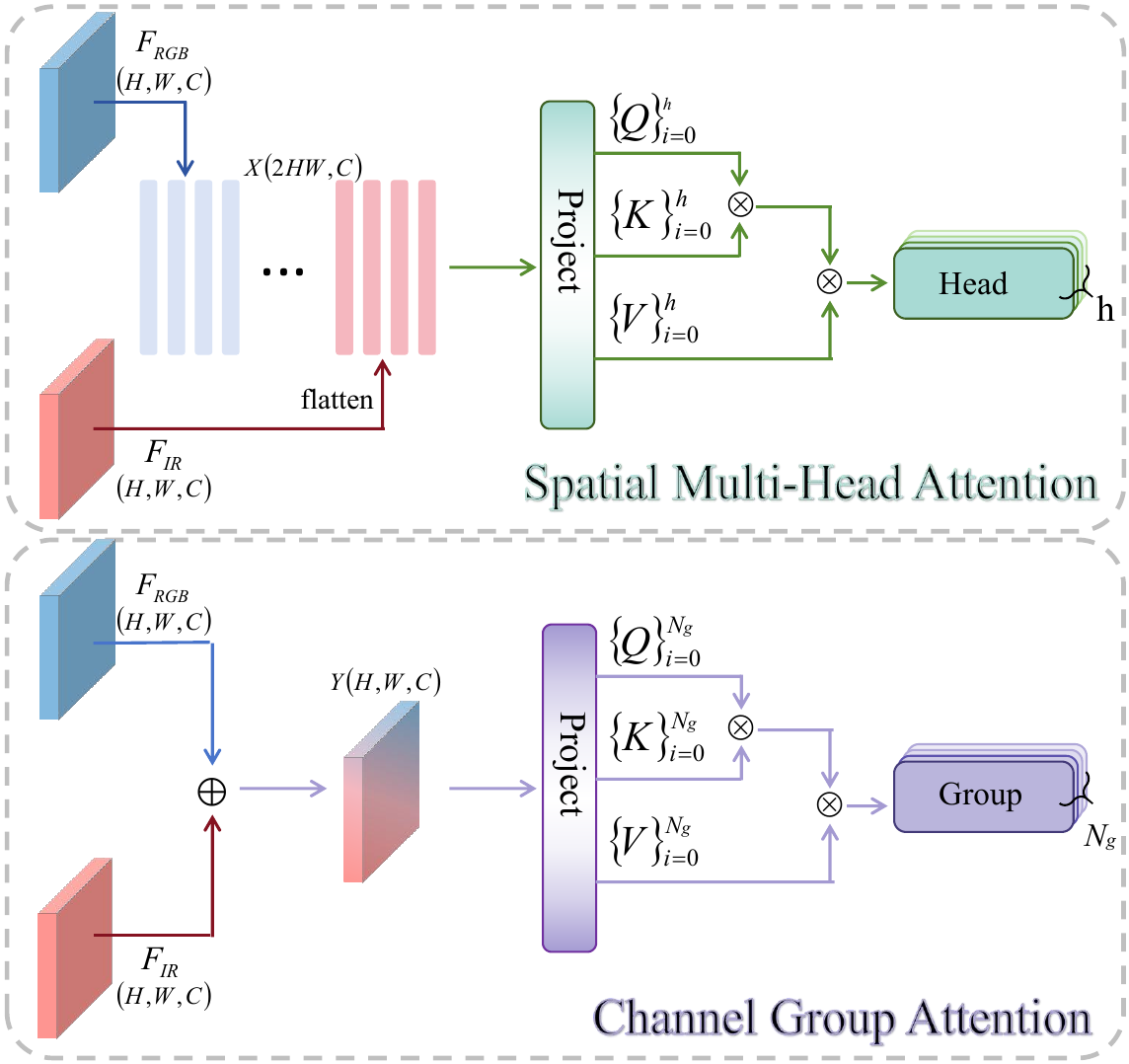}
\caption{Model architecture for our dual attention block. In this work, we use these two types of attention to obtain both local fine-grained and global features.}
\label{fig:3}
\end{figure}

\begin{equation}
Q= XW^{Q} ,
\end{equation}
\begin{equation}
K= XW^{K} ,
\end{equation}
\begin{equation}
V= XW^{V} ,
\end{equation}
where $W^{Q} \in \mathbb{R} ^{C\times C}$, $W^{K} \in \mathbb{R} ^{C\times C}$ and $W^{V} \in \mathbb{R} ^{C\times C}$ represent the respective weight matrices. The spatial attention computation process unfolds as follows:
\begin{equation}
S= Attention\left ( Q,K,V \right )= softmax\left ( \frac{QK^{\top } }{\sqrt{d_{k} } }  \right )  V,
\end{equation}
where $d_{k}$ represents the dimensions of the queries $Q$ and keys $K$, ensuring the inner product remains manageable.

To further refine the self-attention mechanism, our paper introduces the Multi-Head Attention Mechanism, as shown in Fig. \ref{fig:3} $(a)$. Unlike a singular set of $Q$$\backslash$$K$$\backslash$$V$ weight matrices, this mechanism comprises m independent sets, each facilitating the mapping of input vectors to unique representational subspaces. Post multi-head attention processing, we gather multiple weighted matrices $S_{i}$.
\begin{equation}
\begin{split}
S_{i} &= Attention\left ( Q_{i} ,K_{i} ,V_{i}  \right )\\
&= softmax\left ( \frac{Q_{i} K_{i}^{\top } }{\sqrt{d_{k} } }  \right ) V_{i}, i=1,\cdots ,m .
\end{split}
\end{equation}

These matrices are then concatenated and processed through a Linear layer, yielding the spatial MultiHead attention layer's final output $S$.
\begin{equation}
\begin{split}
S&= MultiHeadAttention\left ( X \right ) \\
&= Concat\left ( S_{1},S_{2},\cdots,S_{m} \right ) L,
\end{split}
\end{equation}
where L is the weight matrix for the linear transformation. 

The Transformer further addresses training complexities in multi-layer networks by implementing residual connections, enabling the network to concentrate on the current layer's distinct features.  It also employs layer normalization to equalize the mean and variance of each layer's input, thus expediting the model's convergence.
\begin{equation}
S{}' = LayerNorm\left ( X+  MultiHeadAttention\left ( X \right )  \right ) .
\end{equation}

Following that, $S{}'$, the output from this process, serves as the input to a two-layer fully connected feedforward network.  The first layer utilizes RELU activation functions, while the second layer foregoes activation functions, directly computing the output sequence $S{}''$,
\begin{equation}
S{}'' = LayerNrom\left ( S{}' + FeedForward\left ( S{}' \right )   \right ).
\end{equation}

\subsubsection{Channel Group Attention}
Channel Group Attention, an innovative image-level self-attention mechanism, is tailored to efficiently capture global information, marking a significant departure from earlier approaches. In this mechanism, the channel dimension $C$ determines the token count, while the spatial dimension $HW$ influences the feature size. Each channel token inherently possesses a global nature within the spatial dimension, ensuring that all spatial positions are contemplated during the self-attention process, thereby enhancing the exchange of global information.

While offering several benefits, this method encounters a challenge: the computational complexity escalates considerably with the addition of channel dimensions. To mitigate high computational demands, our solution involves initially merging the feature maps of both modalities through an additive operation, forming the input sentence $Y$. This process preserves the multimodal information without increasing channel dimensionality. Subsequently, we group the feature channels, with $N_{g}$ denoting the number of groups and $C_{g}$ representing the channels per group, thereby defining $C= N_{g} \times C_{g}$. Consequently, the Channel Group Attention mechanism upholds its global scope while enabling effective inter-group information exchange.

Similarly, the input sentence $Y$ is projected onto three weight matrices, yielding sets of queries $Q$, keys $K$, and values $V$,
\begin{equation}
Q= YW^{Q} ,
\end{equation}
\begin{equation}
K= YW^{K} ,
\end{equation}
\begin{equation}
V= YW^{V} ,
\end{equation}
where $W^{Q} \in \mathbb{R} ^{C\times C_{g}}$, $W^{K} \in \mathbb{R} ^{C\times C_{g}}$ and $W^{V} \in \mathbb{R} ^{C\times C_{g}}$ are the weight matrices. The channel attention computation process unfolds as follows:
\begin{equation}
\begin{split}
C_{i} &= Attention\left ( Q_{i} ,K_{i} ,V_{i}  \right )\\
&= softmax\left ( \frac{Q_{i}^{\top } K_{i} }{\sqrt{C_{g} } }  \right ) V_{i}^{\top }, i=1,\cdots ,N_{g} ,
\end{split}
\end{equation}
where we apply attention mechanisms on the transpose of pixel-level tokens to obtain global information, because  each transposed token abstracts the global information.

Then we can obtain the final output C of the channel group attention layer,
\begin{equation}
\begin{split}
C&=ChannelGroupAttention\left ( Y \right ) \\
&=\left \{ Attention\left ( Q_{i} ,K_{i} ,V_{i}  \right )^{\top }   \right \}_{i=0}^{N_{g} }.
\end{split}
\end{equation}

This output, akin to the spatial multihead attention branch, is subsequently processed through a Residual Network and a feedforward network.
\begin{equation}
C{}' = LayerNorm\left ( Y+  ChannelGroupAttention\left ( Y \right )  \right ) .
\end{equation}
\begin{equation}
C{}'' = LayerNrom\left ( C{}' + FeedForward\left ( C{}' \right )   \right ).
\end{equation}

\begin{figure}[htbp]
\centering
\includegraphics[width=0.48\textwidth]{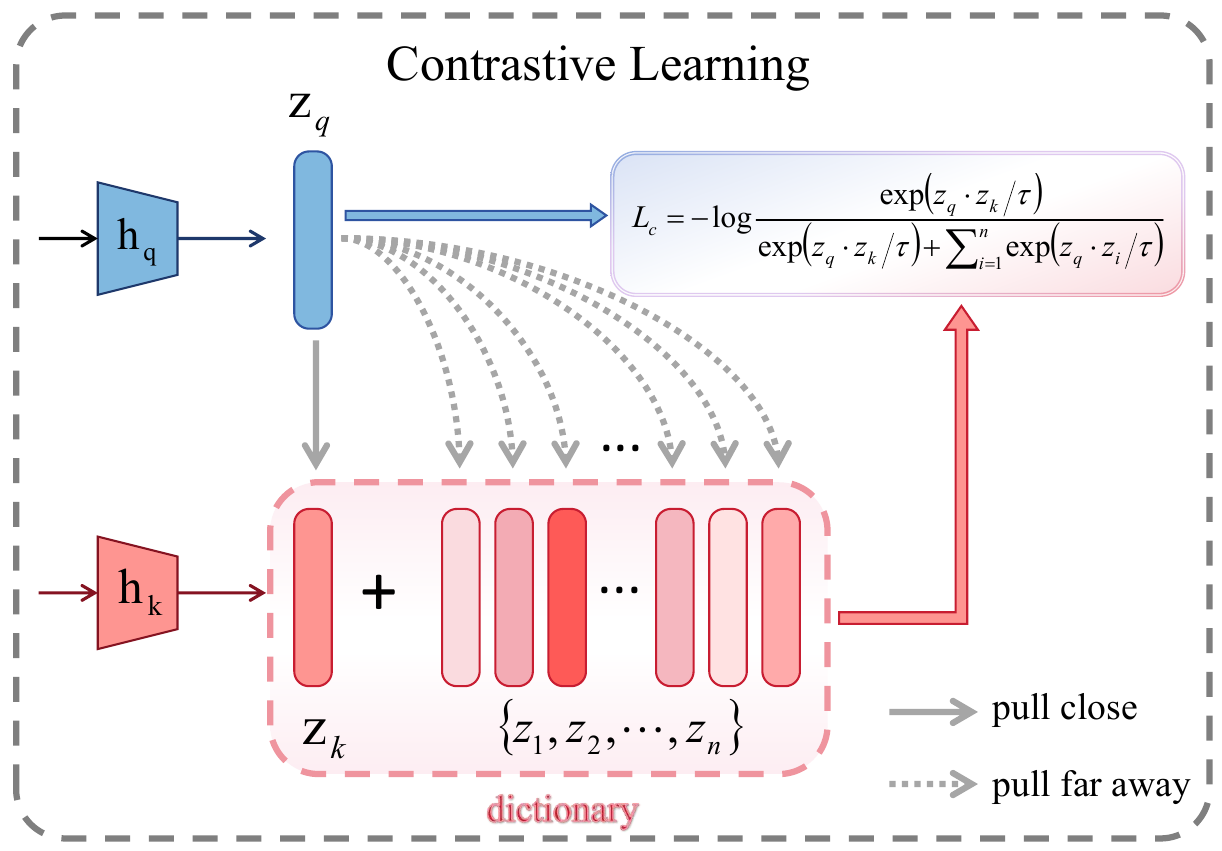}
\caption{Proposed CL module.}
\label{fig:4}
\end{figure}

The outputs from both the Spatial Multi-Head Attention and the Channel Group Attention are added as supplementary information to each modality's respective branch.

The channel group attention mechanism, in contrast to previous methods that relied solely on pixel-level self-attention for information exchange, inherently adopts a global perspective.  This mechanism synergizes with the spatial attention structure: channel group attention provides a global receptive field in the spatial dimension, facilitating feature fusion across global channel tokens to extract high-level global image representations. In contrast, spatial attention refines local representations through fine-grained local interactions across spatial positions, which in turn aids in global information modeling in channel attention.

\subsection{Contrastive Learning} 
\label{Section 3.2}

Contrastive learning effectively learns feature representations by contrasting data points with positive and negative examples in the feature space, tailored for downstream tasks. This learning method shifts focus from intricate pixel details to more abstract semantic information. The depth of features directly correlates with the richness of semantic content, enhancing the efficacy of contrastive learning. Hence, our approach employs contrastive learning on the deepest feature layers, abundant in semantic information, to remedy the Transformer's limitations in extracting the semantic information between visible and infrared image pairs, as shown in Fig. \ref{fig:4}.

Our contrastive learning framework comprises a dual-branch encoder and a projection head. Its core aim is to minimize the distance between a data sample x and its similar positive sample $x^{+}$, while maximizing the gap between x and its dissimilar negative samples $x^{-}$. The main challenge in contrastive learning is constructing these positive and negative samples, while other aspects like the model and loss function are  relatively standardized.

In our study, we use YOLOv7's dual-stream backbone as the encoder to generate queries $q$ and keys $k$ for the contrastive learning network.
\begin{equation}
q= F_{RGB}  ,
\end{equation}
\begin{equation}
k= F_{IR}  .
\end{equation}

We treat each set of multimodal data as different views of the same image processed through diverse augmentations. For features extracted from the RGB modality, we select the corresponding features from the IR modality as positive samples, while features from the other modality serve as negative samples. Subsequently, projection heads $h_{q}$ and $h_{k}$ transform high-dimensional features into representations in the spherical space.
\begin{equation}
z_{q} =h_{q} \left ( q \right )  ,
\end{equation}
\begin{equation}
z_{k} =h_{k} \left ( k \right )  .
\end{equation}

In traditional contrastive learning, the size of the dictionary containing positive and negative samples usually equals the size of the mini-batch. However, the size of the mini-batch is constrained by GPU memory and computational capacity, which cannot be too large, limiting the model's generalization ability. To overcome this, we introduce a queue-based dictionary storage. After each mini-batch is encoded, it enters the queue, and the oldest batch is dequeued. From a consistency perspective, the keys from the earliest calculated mini-batches are the most outdated. In this way, during each iteration update, we can reuse the already encoded keys without needing to update the value of every key element in the dictionary. Meanwhile, the size of the dictionary is no longer limited by the size of the mini-batch but is related only to the queue size. The more keys in the dictionary, the richer the visual information represented, making it easier to find distinctive essential features during matching.

The contrastive learning loss aims to minimize the distance between $z_{q}$ and $z_{k}$, while simultaneously maximizing their divergence from the dictionary queue. This approach essentially frames the problem as an $n+1$ class classification task, where the objective is to classify $z_{q}$ into the $z_{k}$ class.
\begin{equation}
L_{c}= - \log_{}{}\frac{\exp \left ( {z_{q} }\cdot  z_{k}  \right /\tau ) }{\exp\left ( {z_{q} }\cdot  z_{k} /\tau\right ) + \sum_{i=1}^{n} \exp \left ( {z_{q} }\cdot  z_{i} /\tau\right )  }.
\end{equation}

Here $\left \{ z_{k}, z_{1},\dots,z_{n} \right \}$ is the key of a dictionary, where a key matching  $z_{q}$ is written as $z_{k}$, while $\left \{ z_{1},\dots,z_{n} \right \}$ is considered negative keys of  $z_{q}$. $\tau$ is a temperature hyper-parameter.

\subsection{Loss Function}

The overall loss of our network consists of two components:
detection loss $L_{o}$  and contrast learning loss $L_{c}$, which can be
expressed as
\begin{equation}
L=\alpha _{1} L_{o} + \alpha _{2} L_{c} ,
\end{equation}
where $\alpha _{1}$ and $\alpha _{2}$ are weight factors balancing the impacts
of the two losses.

The contrastive learning loss is described above. The detection loss involves three components : loss of judging whether there is an object  $L_{obj}$ , loss of object location$L_{loc}$, and loss of object classification $L_{cls}$, which are used to evaluate the loss of the prediction as
\begin{equation}
L_{o}= \lambda _{obj} \sum_{h=0}^{2} a_{h} L_{obj}+  \lambda _{loc} \sum_{h=0}^{2} b_{h} L_{loc}+  \lambda _{cls} \sum_{h=0}^{2} c_{h} L_{cls},
\end{equation}
where $h$ represents the layer of the output in head, $a_{h}$, $b_{h}$, and $c_{h}$ are the weights of different layers for the three loss functions, the weights $\lambda _{obj}$, $\lambda _{loc}$, and $\lambda _{loc}$ regulate error emphasis among box coordinates, box dimensions, objectness, no-objectness and classification.

\section{Experimental Results}
\label{sec:Experiment}
In this section, the effectiveness of our proposed method is assessed through experiments conducted on the FLIR, LLVIP, and M$^{3}$FD datasets. Initially, we present an overview of the datasets utilized in our study, as shown in Table \ref{tbl:dataset}. Then, we describe the experimental details and evaluation metrics. Finally, we perform a series of ablation studies and compare the proposed method with existing methods to demonstrate the efficacy and advancements of our approach.

\begin{table}[tpb]
	\small
 \renewcommand{\arraystretch}{1.2}
	\centering
	\setlength{\tabcolsep}{3mm}{
		\caption{Dataset Overview}
		\label{tbl:dataset}
		\begin{tabular}{c|ccc}
			\toprule[1.2pt]
			\textbf{Item}  & \textbf{FLIR}  & \textbf{LLVIP} & \textbf{M$^{3}$FD}  \\
			\midrule
			  Classes & 3   & 1  & 6     \\
                Image Size & 640$\times$512 &1280$\times$1024  &1024$\times$768  \\
			Train  & 4129   & 12024  & 3360    \\
			Test & 1013     & 3464   & 840    \\
			Total  &5142  &15488  &4200 \\
                Epoch & 100& 100 & 200\\

			\bottomrule[1.2pt]
	\end{tabular}}
\vspace{-0.1in}
\end{table}
\subsection{Datasets}
\subsubsection{FLIR} 

The FLIR ADAS dataset, a challenging dataset for multimodal object detection, encompasses data captured from both day and night traffic road scenes. A notable challenge within the original dataset is the misalignment of most image pairs, which complicates network training. To address this, our study employs an aligned version of the dataset provided by Zhang \cite{Zhang}, ensuring data consistency. This aligned dataset, based on the original data, comprises 5142 well-aligned visible-infrared image pairs. Of these, 4129 pairs are designated for training and 1013 pairs for testing. The dataset encompasses three distinct object categories: people, car, and bicycle. Throughout this paper, references to the FLIR dataset pertain to this aligned version.
\subsubsection{LLVIP}
The LLVIP dataset is a large-scale visible-infrared image pairing dataset, most of which were collected in very dark scenes to address the poor performance of existing image fusion algorithms under low light conditions. This dataset focuses on pedestrian detection, thus it features a single detection category: pedestrian. It contains 15488 pairs of images, each pair strictly aligned both spatially and temporally, with 12024 pairs designated for training and 3464 pairs for testing.
\begin{table*}[htpb]
	\small
	\centering
	\caption{Ablation Experiments to Validate the Baseline Framework on FLIR Dataset}
	\label{tbl:baseline}
	\scalebox{1.2}{
	\begin{tabular}{c|c|c|c|c|c}
		\toprule[1.2pt]
		\textbf{Model}  & \textbf{Data Type} & \textbf{Method}  & \textbf{$\text{mA}{{\text{P}}_{\text{50}}}$}  & \textbf{$\text{mA}{{\text{P}}_{\text{75}}}$} & \textbf{mAP}\\
		\midrule
		\multirow{4}{*}{Faster R-CNN} & RGB    & Resnet53  & 65.2& 22.1& 29.3 \\
		& IR  & Resnet53  &74.5 &33.0 &37.8\\
            & RGB+IR  & Dual-Stream  & 73.3 &32.4&37.4\\
            & RGB+IR  & Ours   & \textbf{77.8} &\textbf{34.9}&\textbf{39.8}\\
		\midrule
		\multirow{4}{*}{YOLOv3} & RGB   & Darknet53  & 57.9 & 19.7&25.8\\
		& IR & Darknet53    & 73.2   & 31.8 &37.0 \\
            & RGB+IR & Dual-Stream    & 72.5   & 30.7&36.8 \\
            & RGB+IR & Ours    & \textbf{76.8}   & \textbf{32.6} &\textbf{39.1} \\
		\midrule
		\multirow{4}{*}{YOLOv5} & RGB   & CSPDarknet53  & 67.8 & 26.1& 31.2\\
		& IR & CSPDarknet53    & 74.4    & 34.5  &38.0\\
            & RGB+IR & Dual-Stream    & 73.2    & 31.8 & 37.1\\
            & RGB+IR & Ours    & \textbf{79.1}    & \textbf{35.3} &\textbf{40.4} \\
		\midrule
		\multirow{4}{*}{YOLOv7} & RGB   & ELAN-Net  & 68,2 & 26.1 & 31.9\\
		& IR & ELAN-Net    & 75.1    & 33.8  & 38.3\\
            & RGB+IR & Dual-Stream    & 74.9    & 31.6   & 37.6  \\
            & RGB+IR & Ours    & \textbf{80.3}    & \textbf{35.8}  &\textbf{41.3}  \\
		\bottomrule[1.2pt]
	\end{tabular}}
\vspace{-0.1in}
\end{table*}
\subsubsection{M$^{3}$FD}
The M$^{3}$FD dataset is collected using a synchronized imaging system constructed from well-calibrated infrared and optical sensors, covering various scenes including lighting, seasons, and weather conditions with different object types. It annotates 33603 objects across six categories: people, car, bus, lamp, motorcycle, and truck, commonly found in surveillance and autonomous driving scenarios. M$^{3}$FD has amassed a total of 4200 image pairs, with 3360 pairs designated for training and 840 for testing. The dataset's substantial size and rich diversity make it an ideal foundation for learning and evaluating object detection using fused images.

\subsection{Implementation Details}
Our proposed framework is implemented in PyTorch and runs on a workstation with an NVIDIA A100 GPU. In pursuit of better performance, we adopt the YOLOv7 model pre-trained on the COCO dataset as weight initialization. The standard Stochastic Gradient Descent (SGD) is used to train the network with a momentum of 0.98, a weight decay of 0.001, and a batch size of 4. The learning rate is set to 0.01 initially. Meanwhile, we set the temperature scaling parameters $\tau$  at 0.07 in the contrastive loss component. Additionally, the balance weight factors $\alpha _{1}$ and $\alpha _{2}$ in the total loss are set to 1 and 0.1, respectively.
\subsection{Evaluation Metrics}
In this paper, all models are evaluated using three object detection metrics introduced by MS-COCO: mean Average Precision (mAP), $\text{mA}{{\text{P}}_{\text{50}}}$, and $\text{mA}{{\text{P}}_{\text{75}}}$. The mAP is a comprehensive metric obtained by averaging the Average Precision (AP) values across all categories.  It quantifies the area enclosed by the Precision-Recall curve and the axes for all categories using an integral approach. Hence, the mAP can be calculated by
\begin{equation}
mAP= \frac{1}{n} \sum_{i=0}^{n} AP_{i} =\frac{1}{n} \sum_{i=0}^{n}\int_{0}^{1} Precison\ d\left ( Recall \right ),
\end{equation}
where n is the number of categories, the calculations of the
precision and recall metrics are defined as
\begin{equation}
Precison=\frac{TP}{TP+  FP} ,
\end{equation}
\begin{equation}
Recall=\frac{TP}{TP+  FN} .
\end{equation}

TP stands for True Positive, meaning the detector's predicted box and the ground truth (GT) meet the Intersection over Union (IoU) threshold; otherwise, it is considered a False Positive (FP). False Negative (FN) indicates the detector failed to identify a real target. 

$\text{mA}{{\text{P}}_{\text{50}}}$ and $\text{mA}{{\text{P}}_{\text{75}}}$ calculate the average of the AP values for all categories at IoU=0.50 and IoU=0.75 respectively, while mAP can be described as the average mAP over different IoU thresholds (ranging from 0.5 to 0.95 with a step size of 0.05). Clearly, mAP is much stricter than the other two metrics; hence, we consider mAP as the primary measure of difficulty.
\begin{table*}[htpb]
	\small
	\centering
	\caption{Comparison of AP Among the Proposed Method, the Base-detection Method, and Ablation Eeperiment on the FLIR Dataset}
	\label{tbl:ablation}
	\scalebox{1.2}{
	\begin{tabular}{c|c|c|c|c|c}
		\toprule[1.2pt]
        \makebox[0.1\textwidth][c]{\textbf{Dataset}} & \makebox[0.08\textwidth][c]{\textbf{DTF}} & \makebox[0.08\textwidth][c]{\textbf{CL}}&\makebox[0.05\textwidth][c]{\textbf{$\text{mA}{{\text{P}}_{\text{50}}}$}} &\makebox[0.05\textwidth][c]{\textbf{$\text{mA}{{\text{P}}_{\text{75}}}$}}  &\makebox[0.05\textwidth][c]{ \textbf{mAP}}            \\
		\midrule
		\multirow{4}{*}{FLIR} &     &   & 74.9  & 31.6  &37.6 \\
		& $\surd $   &   & 79.7 & 35.2  &40.0\\
            &   & $\surd $  & 78.6 & 33.9  &39.4\\
            & $\surd $  & $\surd $  & \textbf{80.3} \textcolor{blue}{(+5.2)}& \textbf{35.8} \textcolor{blue}{(+3.8)} &\textbf{41.3} \textcolor{blue}{(+3.7)}\\
		\midrule
		\multirow{4}{*}{LLVIP} &     &   &95.8  & 70.9 &63.0 \\
		& $\surd $   &   & 97.5 & 73.2 & 64.3\\
            &   & $\surd $  & 97.2 & 72.8 & 64.7\\
            & $\surd $  & $\surd $  & \textbf{97.9} \textcolor{blue}{(+2.1)} & \textbf{74.3} \textcolor{blue}{(+3.4)} &\textbf{66.1} \textcolor{blue}{(+3.1)}\\
		\midrule
		\multirow{4}{*}{M$^{3}$FD} &     &   & 86.7  & 62.1 & 58.3\\
		& $\surd $   &   & 87.6 & 63.6  & 59.8\\
            &   & $\surd $  & 87.8 & 63.4  & 59.5\\
            & $\surd $  & $\surd $  &\textbf{88.4} \textcolor{blue}{(+1.7)}& \textbf{65.1} \textcolor{blue}{(+3.0)} & \textbf{60.7} \textcolor{blue}{(+2.4)}\\
		\bottomrule[1.2pt]
	\end{tabular}}
\vspace{-0.1in}
\end{table*}
\subsection{Validation of the Baseline Framework}
To verify the overall effectiveness of our proposed method, we integrated the DTF and CL modules with several high-performance classic detectors, including the two-stage detector Faster R-CNN and the one-stage detectors YOLOv3, YOLOv5, and YOLOv7. Experiments were conducted on the FLIR dataset, and the detection performance of these models was evaluated using $\text{mA}{{\text{P}}_{\text{50}}}$, $\text{mA}{{\text{P}}_{\text{75}}}$, and mAP as metrics, with specific results presented in Table \ref{tbl:baseline}. It was observed that all models performed worse in detecting solely with the RGB modality compared to the IR modality alone, primarily due to the presence of numerous low-light scenes in the FLIR dataset, leading to missed detections of valid targets. Moreover, comparing the performance of models using only the IR modality with that of the dual-stream baselines revealed that simple dual-stream networks fail to fully exploit the inherent complementarity between different modalities, sometimes even increasing the difficulty of network learning and resulting in performance degradation.

More importantly, compared to simple dual-stream backbone networks, our method achieved significant improvements in multimodal object detection performance, demonstrating the effectiveness and portability of the DTF and CL modules in multimodal object detection networks. Specifically, when combined with Faster R-CNN, our method improved $\text{mA}{{\text{P}}_{\text{50}}}$ by 4.5$\%$, $\text{mA}{{\text{P}}_{\text{75}}}$ by 2.5$\%$, and mAP by 2.4$\%$. With YOLOv3 and YOLOv5, the enhancements in mAP were 2.3$\%$ and 3.3$\%$, respectively. Our method performed best when integrated with YOLOv7, with performance increases of 5.4$\%$, 4.2$\%$, and 3.7$\%$.

\subsection{Ablation Experiments}
In this section, we mainly evaluate the performance of the proposed method and measure the effects of different module on the network. First, we verify the optimal loss function weight ratio on the FLIR dataset. Then, we conducted ablation experiments on FLIR, LLVIP and M$^{3}$FD datasets to deeply study the contribution of DTF module and CL module to network performance, and analyzed the experimental results in detail.

\subsubsection{Loss function weight factors}
\begin{table}
    \centering
    \setlength{\tabcolsep}{2.8mm}{
		\caption{Ablation Experiment of Loss Function Weight on FLIR Dataset}
		\label{tbl:loss}
        \renewcommand\arraystretch{1.6}
        \begin{tabular}{c|cccc} 
        \toprule[1.2pt]
        $\sfrac{\alpha_{1}}{\alpha_{2}}$       & $\sfrac{1}{0.2}$ & $\sfrac{1}{0.15}$   & $\sfrac{1}{0.1}$   & $\sfrac{1}{0.05}$    \\ 
        \midrule
        $\text{mA}{{\text{P}}_{\text{50}}}$  &79.0 & 79.0&\textbf{80.3} &78.3 \\
        $\text{mA}{{\text{P}}_{\text{75}}}$ &34.9 & 35.1&\textbf{35.8} &33.8\\
        $\text{mAP}$ &40.8 &41.1 & \textbf{41.3} &  40.1   \\
        \bottomrule[1.2pt]
        \end{tabular}
        \vspace{-0.1in}}
\end{table}
To achieve optimal training results for the SeaDATE, we design ablation experiments for the values of $\alpha_{1}$ and $\alpha _{2}$. We maintain the weight of detection loss at $1$, while varying the weight of the contrast learning loss during the training process. We test and evaluate trained models with different weight ratios on the object detection task of FLIR datasets, resulting in Table \ref{tbl:loss}. We find that the weight coefficient ratio is very crucial during the training phase. As shown in Table \ref{tbl:loss}, the best integration effect of object detection network and constrast learning occurs when the weight coefficients are $1$ and $0.1$, respectively. However, when the weight coefficient ratio is too large or too small, it is not conducive to the integration of them.

\subsubsection{Effectiveness of DTF}
In the process of feature extraction in the dual-stream backbone network, we embedded the DTF module multiple times to facilitate the fusion of information from two modalities. The dual-attention mechanism in the DTF module simultaneously considers spatial and channel perspectives. Spatial dimension fusion emphasizes fine-grained local interactions across spatial positions, which is beneficial for local localization tasks. Channel dimension fusion captures global image representations, which is advantageous for classification tasks. To validate the effectiveness of this module, we conducted ablation experiments on the DTF module and evaluated the impact of enabling or disabling the DTF module on the performance of the baseline detectors on the FLIR, LLVIP, and M$^{3}$FD datasets. As shown in Table \ref{tbl:ablation}, on the three multimodal datasets, the DTF module improved the mAP performance of the baseline detectors by 2.4$\%$, 1.3$\%$, and 1.5$\%$, with the remaining evaluation metrics also showing improvement.   These results confirm the effectiveness of the DTF module.

\begin{table*}[htpb]
	\small
	\centering
	\caption{Comparison of Performances on FILR Dataset}
	\label{tbl:comparisionFLIR}
	\scalebox{1.2}{
        
	\begin{tabular}{c|c|c|c|c|c}
		\toprule[1.2pt]
		\textbf{Model}  & \textbf{Data Type} & \textbf{Backbone}  & \textbf{$\text{mA}{{\text{P}}_{\text{50}}}$}  & \textbf{$\text{mA}{{\text{P}}_{\text{75}}}$} & \textbf{mAP}\\
		\midrule
        \multicolumn{6}{c}{mono-modality}\\
        \midrule
        SSD &RGB &VGG16 & 53.1 & 16.3 & 22.6\\
        SSD &IR &VGG16 & 66.2 & 23.2 & 30.4\\
        YOLOv7  &RGB  &ELAN-Net & 68.2 & 26.1 & 31.9\\
        YOLOv7  &IR   &ELAN-Net & 75.1 & 33.8 & 38.3\\
        \midrule
        
        \multicolumn{6}{c}{multi-modality}\\
        \midrule
        Halfway & RGB+IR & VGG16 &71.4 &31.4 & 36.1\\
        GAFF & RGB+IR & ResNet18 & 73.4 &32.6 & 37.6\\
        ProbEn  & RGB+IR & ResNet50 & 75.2 & 32.3 & 38.1\\
        CFT & RGB+IR &CSPDarknet53 & 78.2 &34.8 & 39.6\\ 

        \midrule
            \textbf{SeaDate}(Ours) &RGB+IR & ELAN-Net & \textbf{80.3} & \textbf{35.8} & \textbf{41.3}\\
  
		\bottomrule[1.2pt]
	\end{tabular}}
\vspace{-0.1in}
\end{table*}
\begin{figure*}[htbp]
\centering
\vspace{-2.0em}
\includegraphics[width=0.96\linewidth]{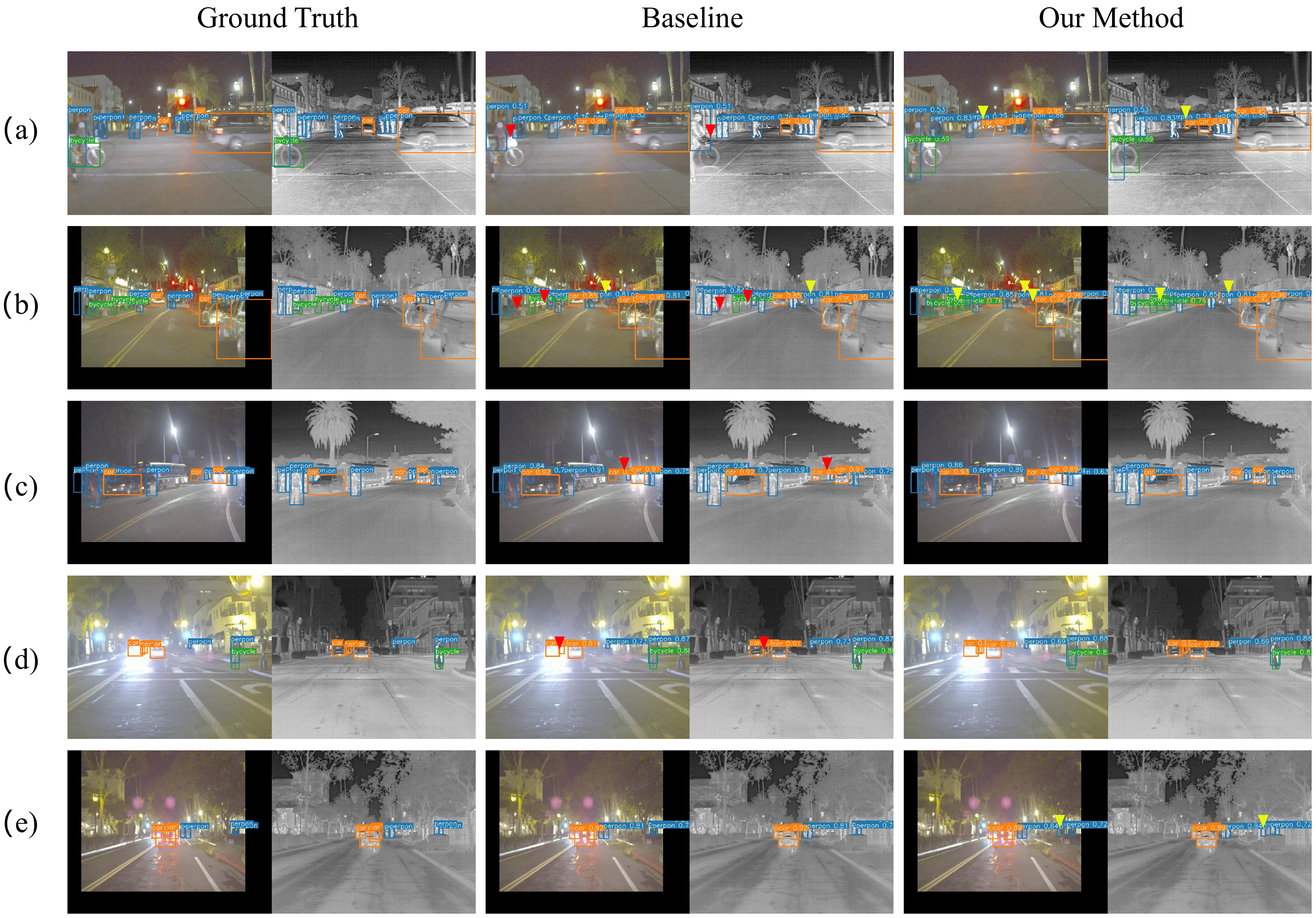}
\caption{Qualitative comparison of multimodal object detection in the FLIR dataset. First column: ground truth, second column: detection results of the baseline, third column: detection results of our method. Note that yellow inverted triangles indicate FPs, and red inverted triangles show FNs. Zoomed in to see details.}
\label{fig:5}
\end{figure*}

\begin{table*}[htpb]
	\small
	\centering
	\caption{Comparison of Performances on LLVIP Dataset}
	\label{tbl:comparisionLLVIP}
	\scalebox{1.2}{
	\begin{tabular}{c|c|c|c|c|c}
		\toprule[1.2pt]
		\textbf{Model}  & \textbf{Data Type} & \textbf{Backbone}  & \textbf{$\text{mA}{{\text{P}}_{\text{50}}}$}  & \textbf{$\text{mA}{{\text{P}}_{\text{75}}}$} & \textbf{mAP}\\
		\midrule
        \multicolumn{6}{c}{mono-modality}\\
        \midrule
        SSD &RGB &VGG16 & 82.7 & 31.5 & 40.2\\
        SSD &IR &VGG16 & 90.6 & 57.1 & 53.8\\
        YOLOv5  &RGB  &CSPDarknet53 & 90.4 & 51.3 & 50.1\\
        YOLOv5  &IR   &CSPDarknet53 & 94.2 & 71.0 & 63.5\\
        \midrule
        
        \multicolumn{6}{c}{multi-modality}\\
        \midrule
        Halfway & RGB+IR & VGG16 &91.4 &60.1 & 55.1\\
        ProbEn  & RGB+IR & ResNet50 & 93.4 & 50.2 & 51.5\\
        
        CFT & RGB+IR &CSPDarknet53 & 97.5 &72.2 & 61.9\\ 
        \midrule
        \textbf{SeaDate}(Ours) &RGB+IR & ELAN-Net & \textbf{97.9} & \textbf{74.3} & \textbf{66.1}\\
  
		\bottomrule[1.2pt]
	\end{tabular}}
\vspace{-0.1in}
\end{table*}
\begin{figure*}[htbp]
\centering
\includegraphics[width=0.96\linewidth]{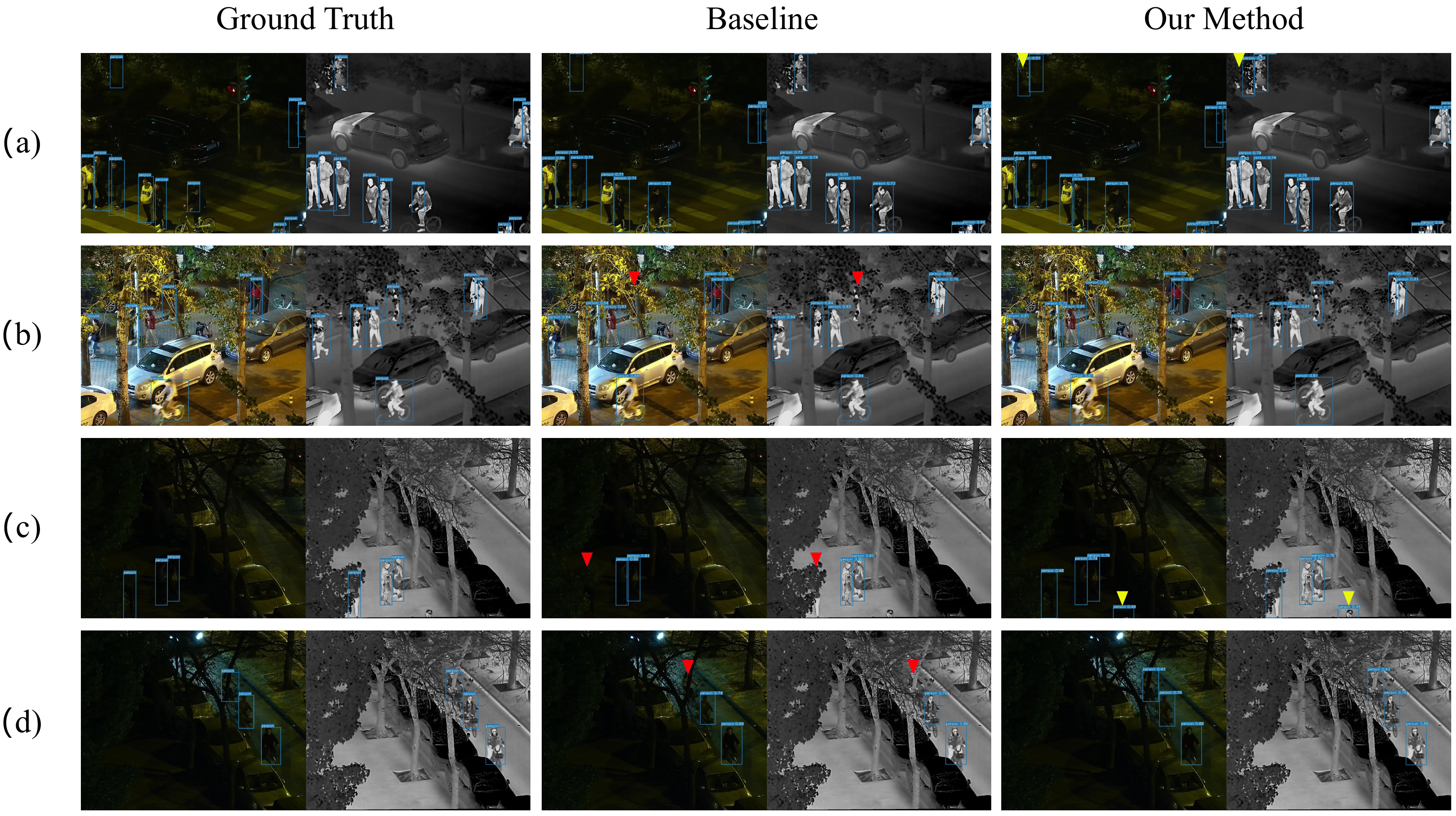}
\caption{Qualitative comparison of multimodal object detection in the LLVIP dataset. Note that yellow inverted triangles indicate FPs, and red inverted triangles show FNs. Zoomed in to see details.}
\label{fig:6}
\end{figure*}

\subsubsection{Effectiveness of CL}
Compared to semantic information on deep features, the Transformer-guided DTF fusion method performs better on detailed information in shallow features. To address this, we employed the CL module. The CL module focuses more on abstract semantic information and learns multimodal features by contrasting data with positive and negative samples in the feature space. Similar to the DTF ablation experiments, we set the same experimental parameters to verify the effectiveness of the CL module on the performance of the baseline detectors. As shown in Table \ref{tbl:ablation}, on the FLIR, LLVIP, and M$^{3}$FD datasets, the CL module improved the mAP performance of the baseline detectors by 1.8$\%$, 1.7$\%$, and 1.2$\%$, with noticeable improvements in $\text{mA}{{\text{P}}_{\text{50}}}$ and $\text{mA}{{\text{P}}_{\text{75}}}$.

In summary, we conducted ablation experiments on each module separately, and the experimental results confirmed the independent effectiveness of the designed modules. Furthermore, the combined use of these modules also highlighted their effectiveness and compatibility. As shown in Table \ref{tbl:ablation}, compared to the baseline detectors, the SeaDATE network's detection performance improved by 3.7$\%$, 3.1$\%$, and 2.4$\%$ on the three datasets, demonstrating higher detection performance than individual modules. This confirms the necessity of the DTF and CL modules and validates the network's strong performance.

\subsection{Comparison With State-of-the-Art Methods}

In this section, we compare the state-of-the-art methods on three datasets, FLIR, LLVIP, and M$^{3}$FD, and show the superiority of the proposed method both qualitatively and quantitatively.

\begin{table*}[htpb]
	\small
	\centering
	\caption{ Comparison of Performances on M$^{3}$FD Dataset}
	\label{tbl:comparisionM3FD}
	\scalebox{1.2}{
	\begin{tabular}{c|c|c|c|c|c}
		\toprule[1.2pt]
		\textbf{Model}  & \textbf{Data Type} & \textbf{Backbone}  & \textbf{$\text{mA}{{\text{P}}_{\text{50}}}$}  & \textbf{$\text{mA}{{\text{P}}_{\text{75}}}$} & \textbf{mAP}\\
		\midrule
        \multicolumn{6}{c}{mono-modality}\\
        \midrule
        YOLOv7  &RGB  &CSPDarknet53 & 81.7 & 55.8 & 50.1\\
        YOLOv7  &IR   &CSPDarknet53   &79.8 & 54.4 & 49.2\\
        \midrule
        
        \multicolumn{6}{c}{multi-modality}\\
        \midrule
        DenseFuse & RGB+IR & CSPDarknet53 &81.6 &56.1 & 50.7\\
        TarDAL  & RGB+IR & CSPDarknet53 & 82.7 & 57.4 & 51.2\\
        
        CFT & RGB+IR &CSPDarknet53 & 84.8 &62.3 & 58.2\\ 
        \midrule
        SeaDate(Ours) &RGB+IR & ELAN-Net & 88.4 &65.1 & 60.7\\
  
		\bottomrule[1.2pt]
	\end{tabular}}
\vspace{-0.1in}
\end{table*}
\begin{figure*}[htbp]
\centering
\includegraphics[width=0.96\linewidth]{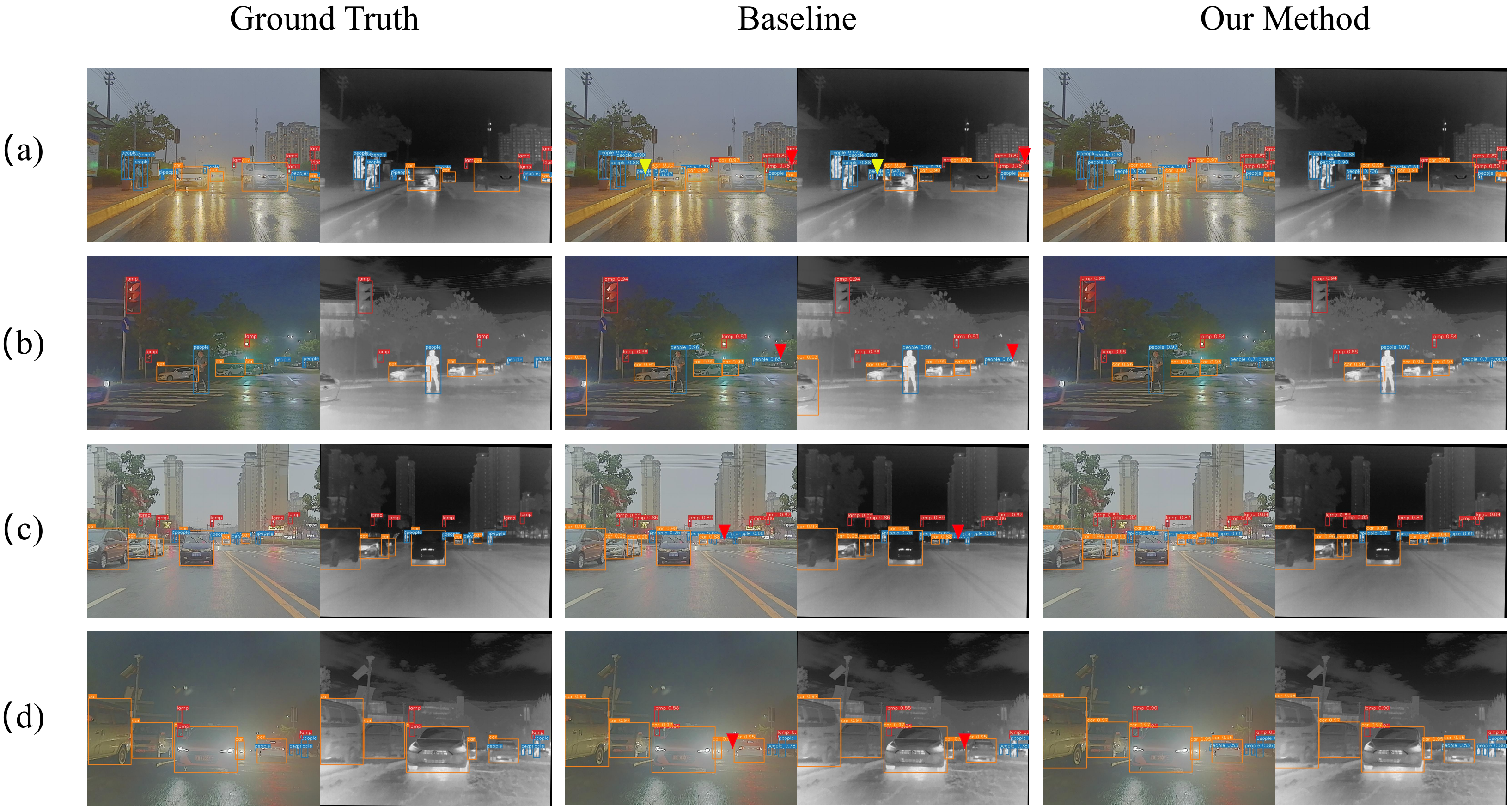}
\caption{Qualitative comparison of multimodal object detection in the M$^{3}$FD dataset. Note that yellow inverted triangles indicate FPs, and red inverted triangles show FNs. Zoomed in to see details.}
\label{fig:7}
\end{figure*}

\subsubsection{On FLIR}
We compare our proposed method with existing methods, including mono-modality detection methods: SSD, YOLOv7, and multi-modality detection methods: Halfway, GAFF, ProbEn, and so on. Table \ref{tbl:comparisionFLIR} shows the detection results of the existing methods and our proposed method on the FLIR dataset. As can be seen, our method achieves state-of-the-art performance on this dataset, with a 3.0$\%$ improvement over the best mono-modality detection algorithms and a 2.1$\%$, 1.0$\%$, and 1.7$\%$ improvement over the best multi-modality detection algorithms on $\text{mA}{{\text{P}}_{\text{50}}}$, $\text{mA}{{\text{P}}_{\text{75}}}$, and mAP, which is an overwhelming performance improvement.

The qualitative detection results on the FLIR dataset are shown in Fig. \ref{fig:5}. The baseline detector using simple additive fusion has a higher rate of missed detections, whereas our method significantly improves detection accuracy by effectively integrating information from two modalities, effectively solving the issue of missed detections. However, during the experiment, we noted that some targets in FLIR dataset were not marked by annotators because they were blocked by trees or other objects, yet our method was able to accurately detect these targets, as shown in Fig. \ref{fig:5} (a) (b) (e), these targets were used as false detection samples, which restricted the improvement of detection accuracy.

\subsubsection{On LLVIP}
The comparison results on the LLVIP dataset are shown in Table \ref{tbl:comparisionLLVIP}. Compared with the best mono-modality object detection method, the detection accuracy of SeaDATE is improved by 2.6$\%$. Compared with the best multi-modality object detection method, the detection accuracy of SeaDATE is increased by 4.2$\%$, which shows the superiority of this method.

From a qualitative analysis perspective, our method excels in addressing challenges such as occlusions and significant modal differences. For example, in Fig. \ref{fig:6} (b), the baseline detector missed detections due to severe occlusion by trees; however, our method significantly enhanced detection performance through improved feature fusion and learning capabilities. Fig. \ref{fig:6} (c) and (d) illustrate the vast differences between modalities, where the lack of information in visible light images severely compromised the effectiveness of infrared images, leading to poor performance by the baseline detector. In contrast, SeaDATE achieved superior detection results through deep interaction between the two modalities' information. Notably, in the LLVIP dataset, as shown in Fig. \ref{fig:6} (a) and (c), our method also detected targets that were not annotated by annotators, similar to the false positive instances observed in the FLIR dataset.

\subsubsection{On M$^{3}$FD}
The comparison results on the M$^{3}$FD dataset are shown in Table \ref{tbl:comparisionM3FD}. Compared with other methods, our method still has the best performance, with an improvement of 11.5$\%$ over the best mono-modality detection algorithm and 2.5$\%$ over the best multi-modality detection algorithm.

The qualitative evaluation of the M$^{3}$FD dataset is shown in Fig. \ref{fig:7}. Compared with baseline detectors, our proposed approach significantly reduces the instances of missed and false detections for small and overlapping targets. By leveraging its robust feature fusion and learning capabilities, SeaDATE effectively enhances target features, thereby improving target detection performance.

\section{Conclusion}
\label{sec:Conclusion}

In this study, we developed SeaDATE, a method for visible-light and infrared target detection. We first designed DTF, a novel fusion technique based on Transformer, employing a dual attention mechanism for interacting spatial and channel-level information to more effectively merge multimodal features. Crucially, our experiments demonstrated that DTF performs better at processing the detailed information of shallow features compared to the semantic information of deep features. Consequently, we introduced the CL module to remedy the shortcomings in semantic information extraction, thus efficiently utilizing cross-modal information. Through the integration of these designs and strategies, SeaDATE achieved superior detection performance across three datasets.





\newpage
{
	\bibliographystyle{IEEEtran}
	\bibliography{reference}
}

\end{document}